%% file: main.tex
\documentclass[10pt,twocolumn,letterpaper]{article}

\usepackage[camera]{cvpr}      % To produce the REVIEW version
\input{preamble}

\definecolor{cvprblue}{rgb}{0.21,0.49,0.74}
\usepackage[pagebackref,breaklinks,colorlinks,citecolor=cvprblue]{hyperref}

\usepackage{graphicx}
\usepackage{float}
\usepackage{caption}
\usepackage{subcaption}

\usepackage{amsmath,bm}
\usepackage{amssymb}
\usepackage{indentfirst}
\usepackage{graphicx}
\usepackage{wrapfig}
\usepackage{color,soul} %\sout

\definecolor{red}{rgb}{1.00,0.00,0.00}
\definecolor{blue}{rgb}{0.00,0.00,1.00}
\definecolor{green}{rgb}{0.30, 0.50,0.00}

\usepackage{flushend} % aligned references columns
\usepackage{multirow}
\def\paperID{16773} % *** Enter the Paper ID here
\def\confName{CVPR}
\def\confYear{2024}

\title{GraspCaps: A Capsule Network Approach for Familiar 6DoF Object Grasping}
\author{Tomas van der Velde$^1$, Hamed Ayoobi$^2$, Hamidreza Kasaei$^{1}$\thanks{Corresponding author: hamidreza.kasaei@rug.nl} \\
$^1$ Department of Artificial Intelligence, University of Groningen, The Netherlands \\
$^2$ Department of Computing, Imperial College London, United Kingdom
}
\begin{document}
\maketitle
\input{sec/0_abstract}    
\input{sec/1_intro}
\input{sec/2_related_works}
\input{sec/3_architecture}
\input{sec/4_dataset}
\input{sec/5_experiments}

\input{sec/conclusion}

\input{sec/ack}
{
    \small
    \bibliographystyle{ieeenat_fullname}
    \bibliography{main}
}


\end{document}

%% file: preamble.tex
\usepackage[dvipsnames]{xcolor}

%% file: sec/0_abstract.tex
\begin{abstract}
As robots become more widely available outside industrial settings, the need for reliable object grasping and manipulation is increasing. In such environments, robots must be able to grasp and manipulate novel objects in various situations. This paper presents GraspCaps, a novel architecture based on Capsule Networks for generating per-point 6D grasp configurations for familiar objects. GraspCaps extracts a rich feature vector of the objects present in the point cloud input, which is then used to generate per-point grasp vectors. This approach allows the network to learn specific grasping strategies for each object category. In addition to GraspCaps, the paper also presents a method for generating a large object-grasping dataset using simulated annealing. The obtained dataset is then used to train the GraspCaps network. Through extensive experiments, we evaluate the performance of the proposed approach, particularly in terms of the success rate of grasping familiar objects in challenging real and simulated scenarios. The experimental results showed that the overall object-grasping performance of the proposed approach is significantly better than the selected baseline. This superior performance highlights the effectiveness of the GraspCaps in achieving successful object grasping across various scenarios.

\end{abstract}

%% file: sec/1_intro.tex
\section{Introduction}
\label{sec:intro}

\begin{figure}[!t]
        \centering
        \includegraphics[width=\linewidth, trim = 0mm 0mm 0mm 0cm, clip=true]{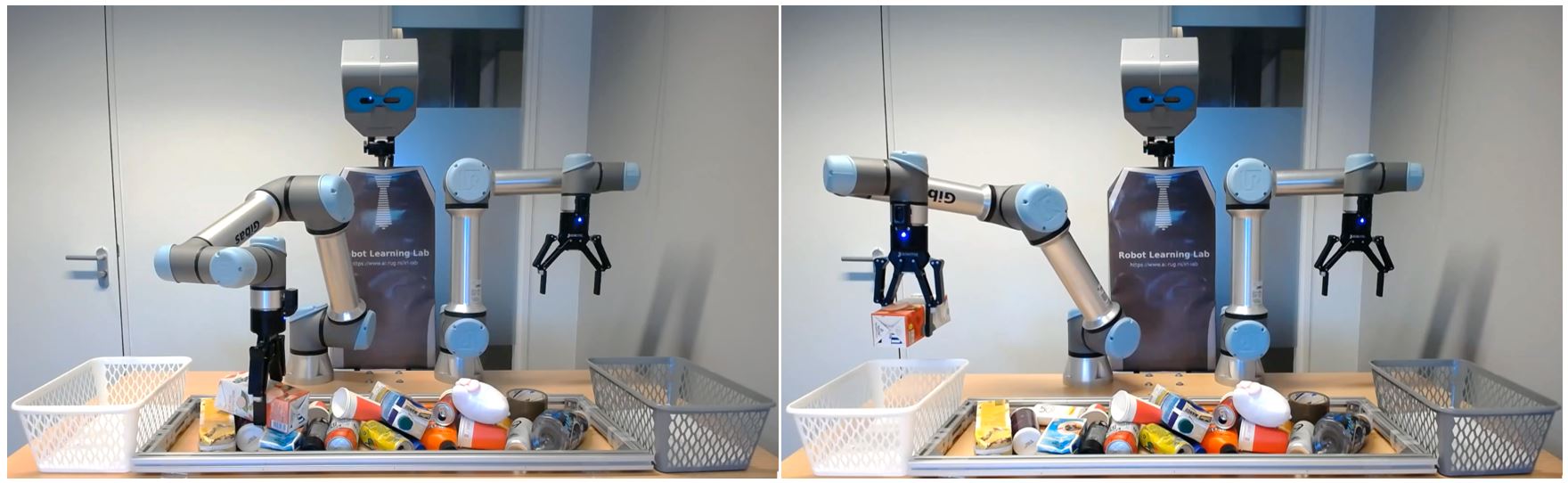}
        \caption{In this illustrative scenario, our dual-arm robot is instructed to do a \textit{clear-table} task. The operational cycle involves processing input point cloud data, predicting a reliable grasp configuration for each point, and subsequently executing the grasping action to transfer the object into the basket. 
        % The successful completion of this task hinges on the robot's ability to accurately predict the grasp configuration for each point of the objects.
        }
        \label{fig:example}
        \vspace{-4mm}
    \end{figure}

Robots are becoming increasingly accessible to the public, finding applications in various non-industrial settings such as homes, hospitals, and shopping malls. This growing accessibility highlights the importance of developing reliable object grasping and manipulation capabilities, as robots must interact with a diverse range of novel objects in dynamic and unforeseen environments (see Fig.~\ref{fig:example}). A significant portion of recent research on object grasping has concentrated on addressing 4 Degrees of Freedom (4DoF) challenges, specifically in achieving object-agnostic grasping. In such approaches, the gripper is typically oriented to approach objects from an overhead perspective (i.e., top-down grasp). However, these methods exhibit notable limitations: (\textit{i}) they do not take into account the semantic function or label of the object, and (\textit{ii}) they inherently restrict the range of interaction possibilities with the object. For example, they cannot distinguish between different objects and struggle to grasp horizontally positioned items like plates. These limitations motivate the study of ``\textit{learning to grasp familiar objects}'', where the robot can recognize the label of the object, and its gripper is free to approach objects from any arbitrary direction it can reach. This approach aims to enhance the versatility and adaptability of robotic grasping in real-world scenarios.

Towards this end, we formulate object grasping as a supervised learning problem based on a capsule network to grasp familiar objects. The primary assumption is that new objects that are geometrically similar to known objects can be grasped in similar ways using object-aware grasping~\cite{kasaei2019interactive, shafii2016learning}. Object-aware grasping allows the network to specifically sample grasps based on the geometry of the object, as opposed to object-agnostic grasping, which generates grasp configurations based on the input to the network without any deeper knowledge of the features of the object it is attempting to grasp. There are several reasons why capsule networks are superior to Convolutional Neural Networks (CNNs) and Vision Transformers (ViT) for grasping objects~\cite{kasaei2021mvgrasp, wang2022transformer}. Their ability to capture spatial hierarchies and part-whole relationships within objects is one of their most notable advantages. Unlike CNNs and ViT, capsule networks are designed to be geometrically invariant, which means they are robust to changes in scale, orientation, and position. Geometric invariance is a crucial property in real-world robotic scenarios where objects are placed in diverse poses. Moreover, capsule networks employ dynamic routing mechanisms, allowing for more flexible information flow between layers. This dynamic routing enables capsules to detect spatial relationships and contributes to a richer understanding of the input data~\cite{sabour2017dynamic}. In the context of object grasping, capsule networks enable object-aware grasping by generating grasp configurations based on object features and geometry. Building upon this concept, we have developed a novel architecture, GraspCaps, that takes as input a point cloud representation of an object and generates as outputs a semantic category label and per-point grasp configurations. Our approach utilizes the activation of a single capsule in the capsule network and processes this activation to produce per-point grasp vectors and corresponding fitness values. To the best of our knowledge, GraspCaps represents the first instance of a grasp network architecture that employs a capsule network for object-aware grasping. The contributions of this paper can be summarized as:

\begin{figure*}[!t]
    \centering
    \includegraphics[width=\textwidth, clip=true]{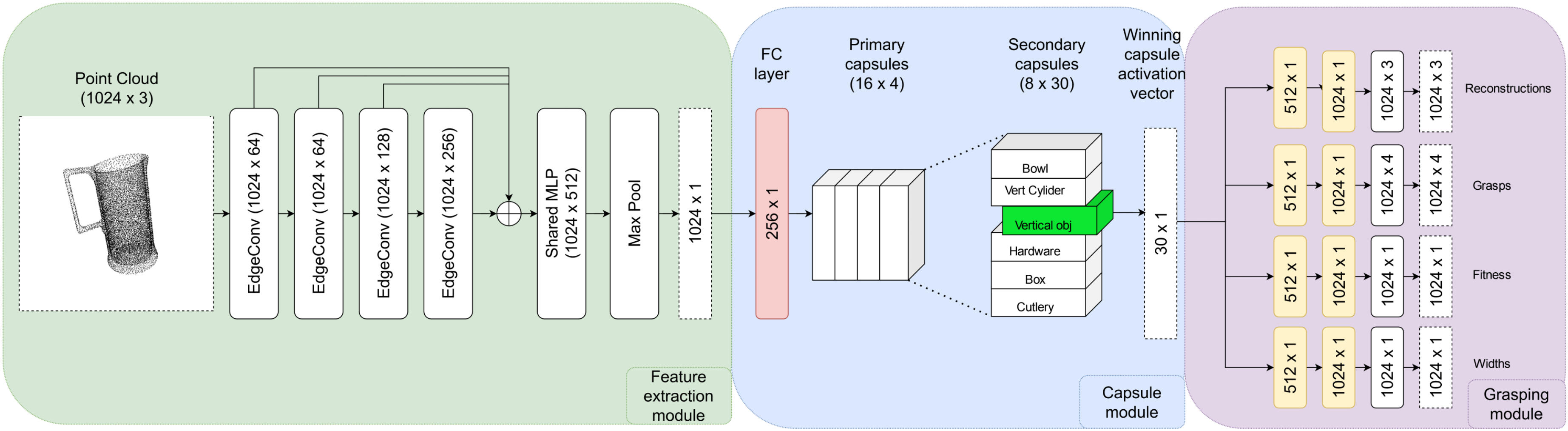}
    \vspace{-5mm}
    \caption{Diagram of the GraspCaps architecture. The architecture is divided into three main modules: the \textit{Feature extraction module}, the \textit{Capsule module}, and the \textit{Grasping module}. 
    The object classification is determined by looking at the activation of each capsule. The capsule with the highest activation is designated as the winner, determining the object present in the input data. Layers depicted in red employ sigmoidal activation, while those in white utilize linear activation. Yellow layers denote fully connected with Leaky ReLU activation.  }
    \label{fig:graspcaps}
     \vspace{-2mm}
\end{figure*}

    \begin{itemize}
        \item This paper presents a novel architecture for object-aware grasping that utilizes a capsule network to process a point cloud representation of an object and generate a corresponding semantic category label along with point-wise grasp synthesis. This marks the first instance of a grasping model using the capsule network.
        \item We propose an algorithm for generating 6D grasp vectors from point clouds and creating a synthetic grasp dataset consisting of $4,576$ samples with corresponding object labels and target grasp vectors. 
        \item To rigorously evaluate the effectiveness of the proposed approach, we conducted a comprehensive series of experiments in both simulation and real-robot setups. 
    \end{itemize}

%===============================================================================

%% file: sec/2_related_works.tex
\section{Related Work}
\label{sec:related}
    % \cred{ I like the content of this section, however, you need to compare the reviewed methods with our approach - what are the main limitation of such approaches and how we address such limitations - mention the similarities and differences between our approach and their works. -- also you can categorize them and mention that our work falls in to which of those categories -- as an example have a look at the related work of https://arxiv.org/pdf/2103.10997.pdf}

    Deep learning-based object grasping methods provide enhanced accuracy and adaptability, reduced dependency on manual engineering, and improved robustness to variability in real-world scenarios~\cite{newbury2023deep}. Current approaches that process point cloud data can be split up into two categories: (\textit{i}) approaches that first transform the point cloud into a different data structure~\cite{santhakumar2022lifelong, kasaei2021mvgrasp, breyer2021volumetric}, (\textit{ii}) approaches that directly process the point cloud~\cite{qi2017pointnet, qi2017pointnet++, li2018pointcnn, shi2020point, wang2019dynamic}. Our method falls into the second category. 
    
    Processing the point set directly has several advantages, since no overhead is added by transforming the point set, and there is no chance of any information loss in the conversion. However, point sets are by definition unordered, which makes extracting local structures and identifying similar regions non-trivial. PointNet~\cite{qi2017pointnet} was one of the first architectures to effectively use point set data for training a neural network in an object recognition task. By design, the PointNet architecture is mostly invariant to point order, which benefits point sets since extracting a natural order from these sets is non-trivial. However, this does limit the performance of PointNet as it cannot recognize local structures in point sets. In prior research the importance of order in data for the performance of neural networks has been illustrated~\cite{vinyals2015ordermatters}, hence order should not be fully disregarded. PointNet++~\cite{qi2017pointnet++} improves upon PointNet by recognizing local structures in the data. Our network architecture is based in part on the architecture used by~\cite{cheraghian20193dcapsule}, which makes the insight to split up the PointNet architecture into several distinct modules.
    
    Later research showed successful results working with point sets by transforming the point set to be processed by a convolutional neural network. PointCNN~\cite{li2018pointcnn} processed the input data by applying a $\chi$-transform on the point set. DGCNN~\cite{wang2019dynamic} and Point-GNN~\cite{shi2020point} employ layer architectures that transform the point set into a graph representation and apply convolution to the resulting graph edges.   
    Several approaches have been successful in processing point sets using a CNN by first transforming the point set into a more regular data structure, such as a 3D voxel grid~\cite{breyer2021volumetric}, top-down view~\cite{santhakumar2022lifelong, kumra2022gr, morrison2018closing}, or multi-view 2D images~\cite{kasaei2021mvgrasp}. The resulting data structures can be processed with existing deep neural network architectures. These conversions come with significant limitations however, as there is a considerable loss in information when converting the point cloud to a different structure, whether that be in the form of losing natural point densities when converting to a voxel grid, or the loss of spatial relations between points when converting to a top-down image. Additionally, the generated voxel grids might be more voluminous than the original point set, as it is likely that many of the voxels remain empty~\cite{qi2017pointnet}. Due to these considerations, we decided to base the GraspCaps architecture in a way that processes the point cloud directly. Moreover, our method utilizes capsule activations to generate per-point grasp configuration. The intricate understanding of spatial hierarchies afforded by capsule networks distinguishes GraspCaps as a pioneering solution for object grasping. Unlike the reviewed approaches, our approach alleviates the need for excessive pooling layers employed in CNN architectures. Such pooling layers can result in a loss of detailed spatial information. 

    In the field of grasp generation, S$^4$G \cite{qin2020s4g} extended the PointNet architecture to generate 6D grasps based on the input point set. Grasp pose detection (GPD)~\cite{ten2017gpd} was developed to generate and evaluate the fitness of grasps. It takes a point cloud as its input and generates grasps which are then filtered on fitness. The network then classifies the grasp candidate as either successful or unsuccessful. PointNetGPD \cite{liang2019pointnetgpd} builds upon the idea of GPD and expands on it by employing the PointNet architecture to evaluate the fitness of grasps. GraspNet~\cite{mousavian2019graspnet} uses a variable autoencoder to generate a set of grasps for an object using a point cloud. It evaluates the fitness of the generated grasps using an encoder-decoder network. It can generate grasps with a success rate comparable to PointNetGPD. In contrast to GPD-based approaches, our method generates point-wise grasp configurations and can classify familiar objects.

%===============================================================================

%% file: sec/3_architecture.tex
\section{GraspCaps}
\label{sec:architecture}

    % The GraspCaps architecture is a combination of several existing architectures. We opted for a modular approach to network design, as this made it easy to quickly test variations on modules, and provides a good start for potential future research. A diagram of the architecture is shown in Fig.~\ref{fig:graspcaps}.
    
    As depicted in Fig.~\ref{fig:graspcaps}, the GraspCaps architecture consists of three main modules: the feature extraction module, the capsule module, and the grasp synthesis module. 
    
    The \textit{\textbf{Feature Extraction Module}} takes a normalized $1024 \times 3$ dimensional point cloud as its input and processes the point cloud using the feature extraction module from the Dynamic Graph CNN (DGCNN)~\cite{wang2019dynamic} to generate a $1024 \times 1$ dimensional feature vector.  DGCNN was chosen for this module due to its ability to capture both local and global context, handle permutation invariance, adapt to irregular sampling, and efficiently learn local structures \cite{cheraghian20193dcapsule}.  Specifically, it processes a point cloud by use of four EdgeConv layers. An EdgeConv layer first transforms the point cloud into $N$ undirected local graphs of $k$ nodes, where $N$ is the number of points in the input point cloud, and $k$ is an integer representing the number of closest neighboring points that are used for constructing the local graph. The network then applies a shared multi-layer perceptron (MLP) on the edges of the graphs. This process is repeated for each of the four EdgeConv layers in the feature extraction module. The output of all five EdgeConv layers is concatenated and processed using a shared MLP and pooled to a $1024 \times 1$ feature vector using an adaptive max-pool operation.

    The \textit{\textbf{Capsule Module}} of our network consists of a single fully connected layer, a primary capsule layer, and a secondary capsule layer. The fully connected layer consists of 256 neurons using sigmoidal activation that transforms the feature vector for use in the primary capsule layer.  The primary capsule layer consists of 16 capsules, each containing 4 neurons. The secondary capsule layer consists of 8 capsules each containing 30 neurons, with each capsule corresponding to one specific shape class of object. The primary and secondary capsule layers are connected through routing by agreement~\cite{sabour2017dynamic}. Capsules in deeper layers encode more meaningful features. Our network is trained to ensure that each capsule in the deepest layer corresponds to a distinct object class. To determine the object class, the Euclidean norms of the activations in the secondary capsule layer are ranked. The capsule with the highest norm is considered the winner, and its activation is used for further processing in the subsequent module. The activations of all other capsules are masked off, preventing them from influencing downstream processing.

    The \textit{\textbf{Grasping Module}} consists of four fully-connected heads. Their architecture is based on the architecture of the reconstruction network as described by~\cite{sabour2017dynamic}, each containing three fully connected layers. The first two layers consist of 512 and 1024 neurons respectively, using leaky ReLU activation. The third layer size is dependent on the output shape of the network and uses linear activation. The reconstruction head has an output shape of $1024 \times 3$ to generate point locations in 3D space. The grasp head has an output shape of $1024 \times 4$ to generate quaternion rotation vectors for each point. The quality and width heads both have an output shape of $1024 \times 1$, as they only need to output a scalar per point. The output of the grasping head is normalized to obtain a unit quaternion vector.

The use of a modular architecture in our design significantly boosts the flexibility and agility of our experimentation process. By breaking down the GraspCaps architecture into distinct modules, we create a structure that facilitates easy modification of individual components. Furthermore, the decision-making process for selecting parameters was thorough and systematic. We conducted extensive parameter sweeps to identify the configurations that optimize the performance of each module (see Sec.~\ref{ablation_studies}). 
% This approach ensures that our model is well-adapted to the task at hand (see Sec.~\ref{ablation_studies}). 
    
\subsection{Loss functions}
    As the network performs several tasks at once, a custom loss function was developed. Our loss function can be broken down into three main components: Margin loss ($\ell_\text{margin}$)~\cite{sabour2017dynamic}, Reconstruction loss ($\ell_\text{recon}$)~\cite{sabour2017dynamic}, and Grasping loss ($\ell_\text{grasp}$). 

    The \textit{\textbf{margin loss}} component, inspired by dynamic routing mechanisms~\cite{sabour2017dynamic}, aims to enhance the recognition capability of the network. It enforces a margin between the activation of the correct capsule and other capsules, ensuring that the network accurately assigns significance to the relevant capsule and minimizes interference from irrelevant ones. In other words,     Margin loss is used to train the capsule activation to be representative of the class it encapsulates:  
    \begin{equation} \label{eq:marginloss}
        \begin{aligned}
            \ell_\text{margin}(\mathbf{v}_i) = & T_i \max(0, m^+ - ||\mathbf{v}_i||)^2 + \\
                                            & \lambda (1-T_i)\max(0, ||\mathbf{v}_i|| - m^-)^2
        \end{aligned}
    \end{equation}
    
    \noindent where $T_i$ is the binary indicator that is introduced to signify the presence of an instance from class $i$ in the input data. Specifically, $T_i = 1$ if an instance from class $i$ exists in the input data. $m^+$ and $m^-$ are constants and set to $0.9$ and $0.1$ respectively. The $\lambda$ used is a scaling factor that is set to $0.5$. It is used to scale down the shrinking of the activity vectors in the event that the output of the capsule is incorrect, which stops the initial learning from shrinking the activation vectors~\cite{sabour2017dynamic}.  The activation vector $\mathbf{v}_i$ corresponds to capsule $i$. These vectors are central to the capsule network's functioning and contain information relevant to the presence of specific classes in the input data.
    By calculating the Euclidean norm, $||\mathbf{v}_i||$, we obtain a scalar that can be treated as the probability that the corresponding class is present in the input.%, similar to the output of more conventional neural networks.
    
    The \textit{\textbf{reconstruction loss}} serves a dual purpose within our architecture, acting as both a regularizer and a mechanism to ensure that each capsule learns information pertinent to its associated object class. Specifically, in our design, the reconstruction loss is formulated as the mean squared error between the input point set and the reconstructed point set. This loss term is then scaled down by a factor $\beta$ to prevent it from dominating other losses, allowing it to function primarily as a regularizer. 
    
    The \textbf{\textit{grasping loss}}, inspired by the framework introduced in~\cite{breyer2021volumetric}, encompasses distinct loss functions for rotation, grasp quality, and grasp width. An enhancement made to this loss function involves the incorporation of the binary indicator $T_i$, mirroring its usage in the margin loss. This inclusion ensures that the grasping network focuses exclusively on capsule activations corresponding to the correct object, preventing the training process from being influenced by irrelevant capsule activations. By incorporating $T_i$ into the grasping loss, we align the training process more closely with the task's objectives, enhancing the network's ability to generate accurate and contextually relevant grasp configurations.
    The second alteration from the loss function is the definition of the quality loss $\ell_\text{quality}(q_i, \hat{q}_i)$, which we extend to allow any real-valued ground truth value $q_i \in [0,1]$ instead of the binary ground truth $q_i \in \{0,1\}$ used by~\cite{breyer2021volumetric}. By this alteration, the network can use $q_i$ like a fitness value and distinguish usable but imperfect grasps from the best-fitting grasps. Bad or colliding grasps are heavily penalized during data generation, so their values are relatively small, which reduces the risk of the network training on bad grasps. The grasping loss is defined as:
   \begin{equation} \label{eq:grasploss}
       \begin{aligned}
           \ell_\text{grasp}(\mathbf{g_i}, \mathbf{\hat{g}}_i) = &T_i [ \ell_\text{quality}(q_i, \hat{q}_i) + \\
                                                                & q_i (\ell_\text{rotation}(\mathbf{r}_i, \mathbf{\hat{r}}_i) + \alpha \ell_\text{width}(w_i, \hat{w}_i))]
       \end{aligned}
   \end{equation}
    \noindent where both $\ell_\text{quality}$ and $\ell_\text{width}$ are defined as the mean squared error loss. %, as found in equation (\ref{eq:mseloss}). 
    The rotation loss is defined in Equation (\ref{eq:rotationloss}), and $\alpha$ is a constant set to $0.001$. As the end-effector of the robotic arm we are using is a symmetrical two-fingered gripper, a grasp vector rotated by $180^{\circ}$ around the gripper wrist results in effectively the same grasp~\cite{breyer2021volumetric}. Therefore, we define the rotational loss function to consider both grasps correct:
   \begin{equation} \label{eq:rotationloss}
       \ell_\text{rotation}(\mathbf{r}, \mathbf{\hat{r}}) = \min(\ell_\text{quat}(\mathbf{r}, \mathbf{\hat{r}}), \ell_\text{quat}(\mathbf{r}\pi, \mathbf{\hat{r}})) 
   \end{equation}
   \noindent where  $\ell_\text{quat}(\mathbf{r}, \mathbf{\hat{r}}) = 1 - |\mathbf{r} \cdot \mathbf{\hat{r}}|$ is defined to use the inner product to calculate the distance between the ground truth rotation $\mathbf{r}$ and the generated rotation $\mathbf{\hat{r}}$.
   The loss function that is used for training the network incorporates all of these loss functions as follows:
   \begin{equation} \label{eq:totalloss}
        \mathcal{L}(\mathbf{t}_i, \mathbf{\hat{t}}_i) = \ell_\text{margin}(\mathbf{v}_i) + \beta \ell_\text{recon}(\mathbf{p}_i, \mathbf{\hat{p}}_i) + \ell_\text{grasp}(\mathbf{g_i}, \mathbf{\hat{g}}_i)
   \end{equation}
  \noindent where $\mathbf{t}_i$ is the output of the network, consisting of the capsule output $\mathbf{v}_i$, the reconstruction of the input point set $\mathbf{p}_i$, and the generated grasp $\mathbf{g}_i$. $\mathbf{\hat{t}}_i$ is the target vector, composed of the input point set $\mathbf{\hat{p}}_i$, and the target grasp $\mathbf{\hat{g}}_i$. $\beta$ is a scaling constant set to $0.0005$.

%===============================================================================

%% file: sec/4_dataset.tex
\section{Generating Grasp Synthesis}
\label{sec:dataset}

Given the specialized nature of our network, which entails simultaneous recognition and grasping on a point cloud, no existing benchmark datasets were identified that precisely met our criteria. Consequently, we generate our own dataset to train and evaluate the proposed approach. A crucial aspect of this dataset creation involved the implementation of an algorithm capable of generating valid grasp configurations based on point cloud data, ensuring the availability of a substantial dataset with accurate grasp targets.

\begin{figure}[!t]
     \centering
     \includegraphics[width =\linewidth]{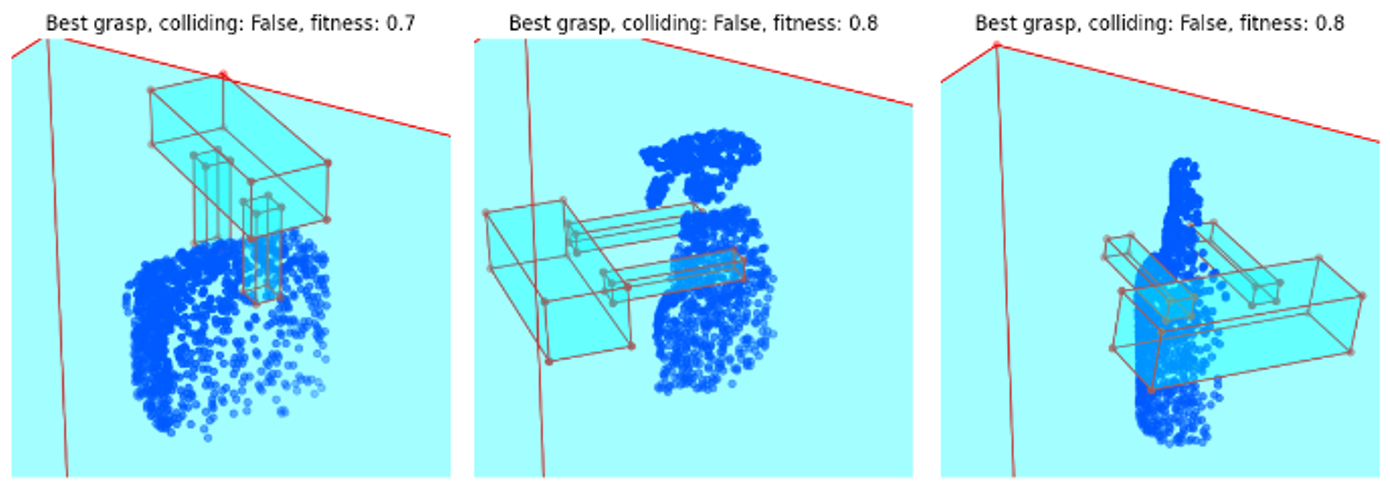} 
    \vspace{-5mm}
    \caption{ During the dataset generation process, only the orientation and opening width of the gripper are altered. The location of the grasp center is determined by choosing a random point in the point cloud before converging using simulated annealing.}
     \vspace{-5mm}
    \label{fig:grasp_gen_main}
\end{figure}
In this work, the gripper is simulated as a combination of three separate boxes: two for the fingers, and one for the base (see Fig.~\ref{fig:grasp_gen_main}). The object placement process begins by randomly positioning an object within the robot's workspace, followed by capturing the point cloud of the object. Subsequently, the point cloud is converted into a watertight mesh to streamline collision-checking efficiency. It's noteworthy that while the original point cloud is retained for grasp fitness calculations, the conversion to a mesh notably accelerates the algorithm by reducing the number of required collision checks per iteration. The algorithm initiates the grasp with a random rotation, presenting the gripper either facing downward towards the object or oriented from the side. Following initialization, the gripper's orientation and opening width undergo iterative updates to converge on a well-fitted grasp. Acceptance of a new state is contingent upon the grasp exhibiting higher fitness than the current state. Alternatively, the new state may be accepted randomly using simulated annealing~\cite{kirkpatrick1983optimization}, introducing a stochastic element to the optimization process. 
     
In simulated annealing, the algorithm starts out with a high temperature. This temperature corresponds with a high chance of accepting new states, even if they have a lower fitness than the current state. The temperature of the system linearly decreases during execution and worse states are less and less likely to be accepted. In the final iterations simulated annealing behaves almost identical to stochastic gradient descent. The high temperature at the start of the process allows the algorithm to escape local maxima and converge to the global optimal solution, while the low temperature at the end of the process ensures that the algorithm converges. In terms of grasp generation, it allows the system to escape usable but imperfect grasps and reach better grasps. The only state transition that is not able to occur even when accepting random state transitions is to go from a non-colliding state to a colliding state. We have also considered bounds on the rotation to ensure the algorithm does not generate grasps that are difficult for the robot to reach. 
    
The fitness of each state is determined by five factors. The first factor considers what percentage of points is located between the two fingers of the gripper. This should increase the likelihood that the algorithm converges on grasping a large part of the object. The second factor determines the minimum distance between a finger and the points between the fingers. It does this for each finger. The third factor looks at how closely the normals of the fingers of the gripper overlap with the normals of the object. If the normals are completely opposed to each other the gripper is perpendicular to the object, and the grasp should have a high chance of success. The fourth factor is defined as the distance between the two grippers. Since the grasp should be as tight as possible, a small penalty is added to the fitness for the distance between the two fingers. The last factor adds a base fitness based on the distance of the grasping point to the center of the object. The closer the point is to the center of the object, the higher its score. Examples of grasp synthesis for different objects are shown in Fig.~\ref{fig:grasp_gen_main}.
    
    %===============================================================================

%% file: sec/5_experiments.tex
\section{Experimental Results}
\label{sec:result}

\subsection{Data Generation and Statistics}
    \begin{figure*}[!t]
       \vspace{-5mm}
       \begin{tabular}{ccc}
          \includegraphics[width=0.325\linewidth]{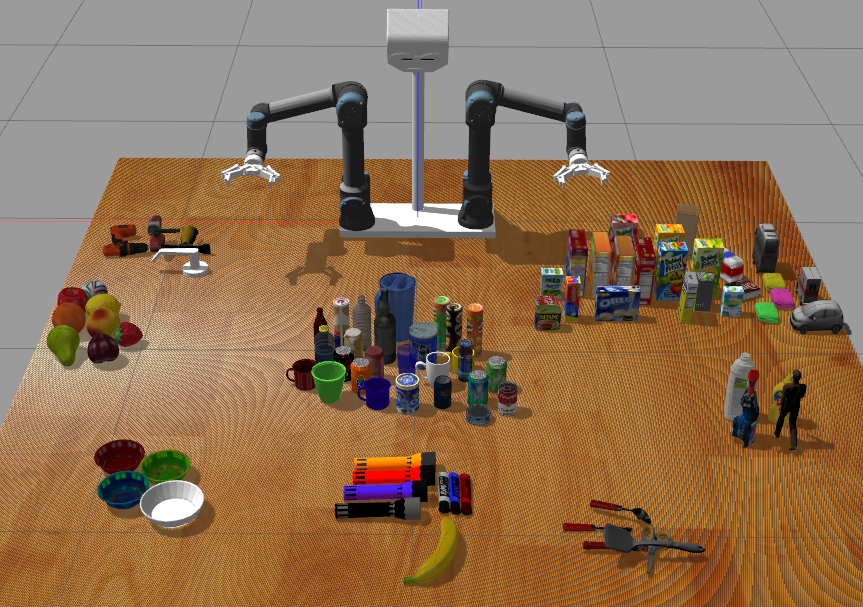}&  
          \includegraphics[width=0.31\linewidth]{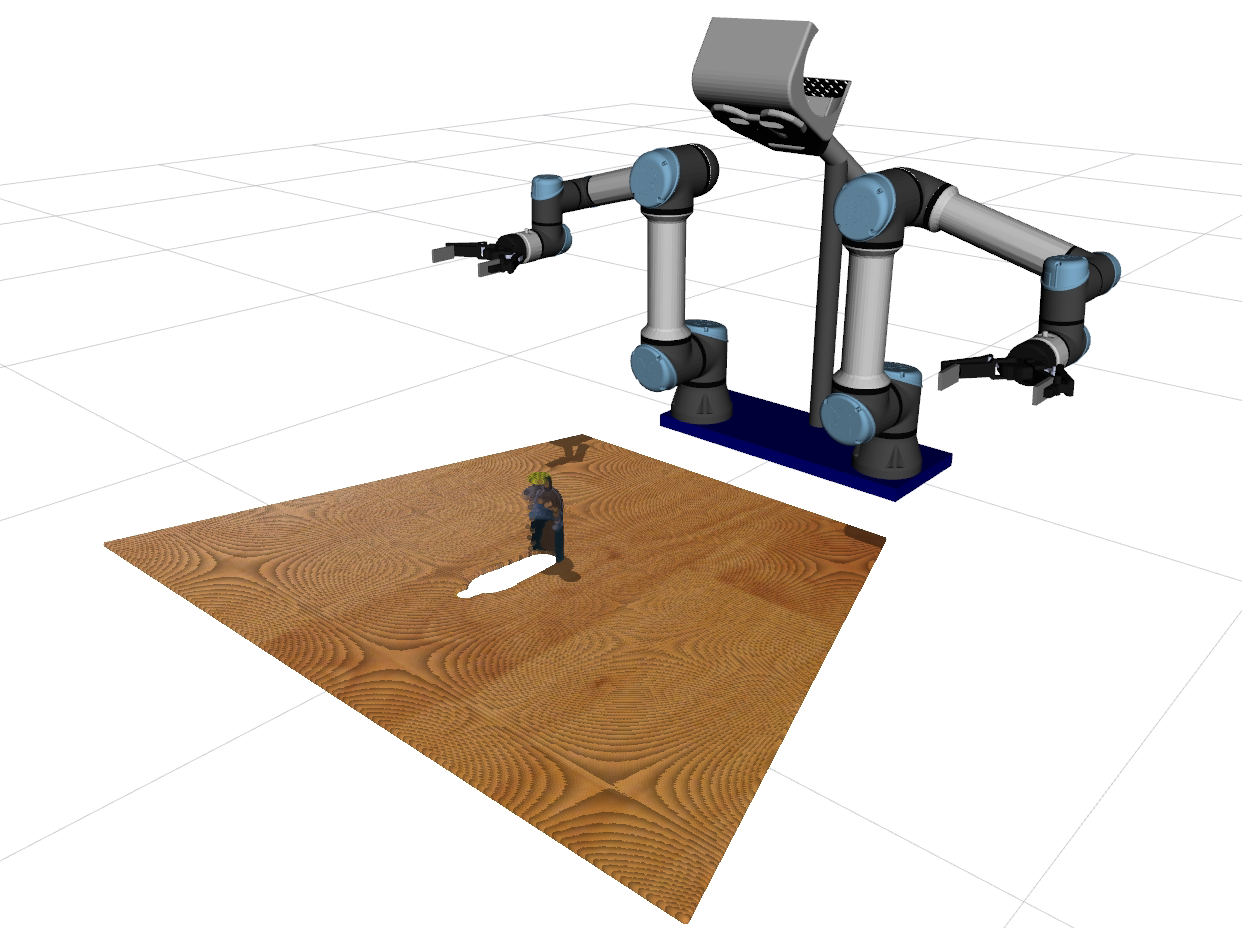}
          \includegraphics[width=0.30\linewidth]{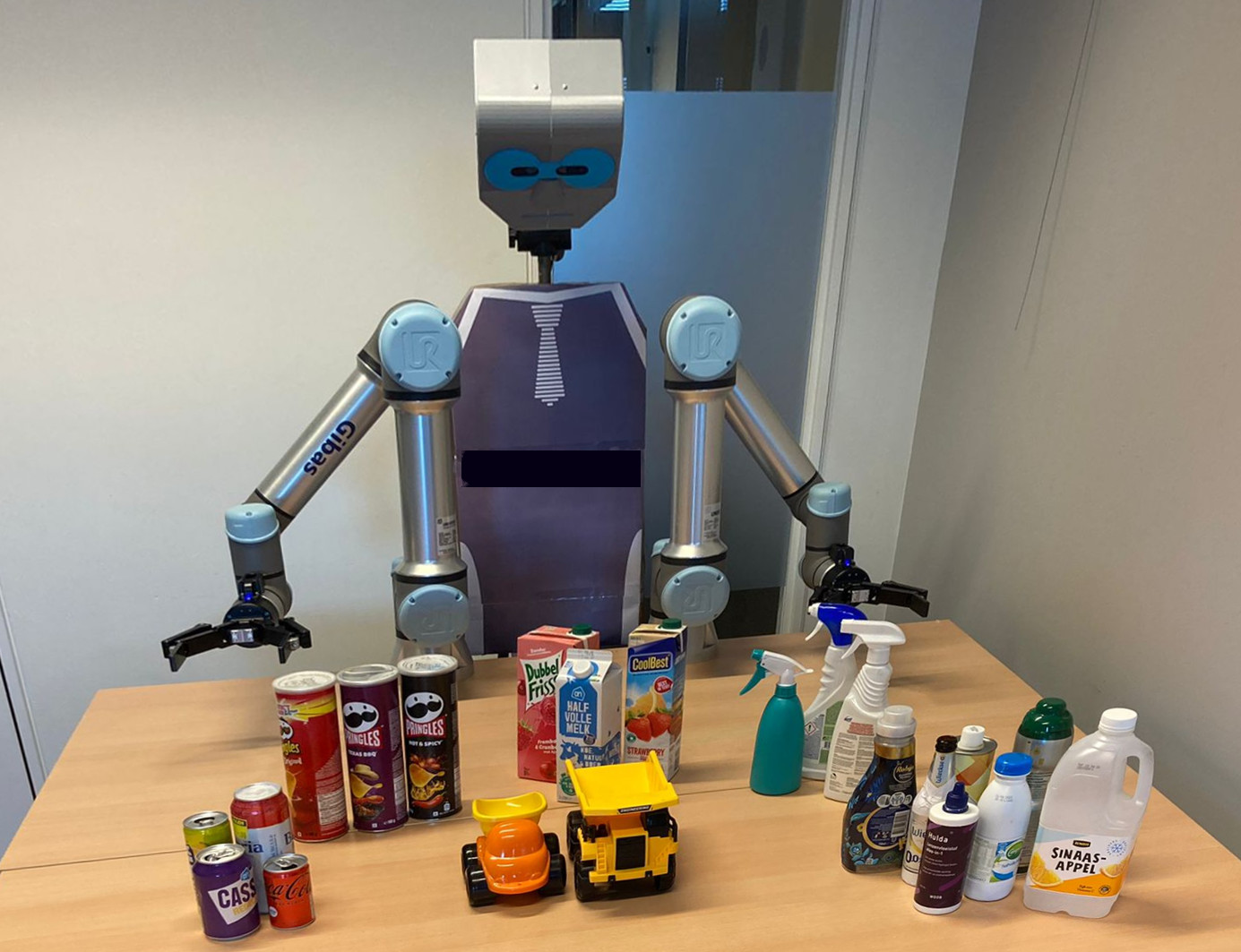}
       \end{tabular}
        \vspace{-3mm}
        \caption{Our experimental setup in simulation and real robot: (\textit{left}) $90$ simulated 3D objects semantically grouped in the simulated Gazebo environment;  (\textit{center}) We randomly place a bottle object in front of the robot and visualize what robot can perceive through its single camera; (\textit{right}) The real robot setup as used for the experiments. All objects placed in front of the robot are used to test the network.}
         \vspace{-3mm}
       \label{fig:all_objs}
    \end{figure*}

\noindent\textbf{Data Generation: }
As shown in Fig.~\ref{fig:all_objs} (\textit{left}), we used $90$ simulated household objects to generate a large synthetic grasp dataset~\cite{kasaei2022lifelong}. Furthermore, to address the familiar object recognition task, we grouped all objects into eight distinct object categories based on shape similarities. For instance, milk cartons are grouped with juice cartons and other boxes. 
Efficient training of the grasp-generation module of the network is facilitated by generating $120$ grasp points per object. These grasp points are centered on random points within the point cloud of the object. Each generated grasp configuration includes information about rotation in quaternion format, a quality measurement, and the opening width of the gripper. To enhance the training process, two distinct datasets are generated. The network is sequentially trained on these datasets, first focusing on the classification section and subsequently on the grasping section. This approach, based on experimentation, has been found to yield superior performance compared to training the network in an end-to-end manner on a single dataset. This sequential training strategy enhances the learning process, allowing the network to progressively refine its understanding of object recognition and grasp synthesis.

\noindent\textbf{Statistics:}  All samples are made by placing an object with random orientation and location. In the case of the synthetic dataset, we randomly placed the object in Gazebo simulation~\cite{koenig2004design} and employed a single sensor measurement using an ASUS Xiton RGBD camera to obtain a point cloud of the object (see Fig.~\ref{fig:all_objs} (\textit{center}). We ensured that the dataset does not become biased towards the classes with a large number of objects such as the box class. 
The obtained synthetic dataset is balanced, consisting of $4,576$ samples, with $572$ samples per class. Moreover, a total of $34,067$ grasp configurations were generated for training the network. 
To evaluate the performance of our approach in real-world settings, a real object dataset was generated using objects corresponding to the classes used in the simulation. This dataset encompasses $220$ unique point clouds of 22 objects, and for each point cloud, we generated $120$ grasp configurations, totaling $26,643$ grasps. Our real object dataset is utilized for fine-tuning the grasping part of the GraspCaps to enhance its performance in real-world conditions. Some of the objects used for creating the dataset are shown in
Fig.~\ref{fig:all_objs} (\textit{right}). In both real and synthetic datasets, we considered $70\%$ of the available data for the training, and the remaining $30\%$ for the validation.

\subsection{Ablation studies}
\label{ablation_studies}
    Through a series of ablation experiments, we investigate the impact the capsule module has on familiar object recognition and grasping. As a baseline, we considered a network based on the GraspCaps architecture, but lacking the capsule layer. This allows the network to generate object-agnostic grasps and allows us to observe whether capsule networks are a viable method of capturing the information needed to generate usable object-aware grasps. The classification module is based on the architecture described in~\cite{wang2019dynamic}. Both networks were first trained on the synthetic dataset for 300 epochs, after which they had their weights frozen in the feature extraction and classification modules. we then trained them on the $120$ grasp per sample dataset for an additional $200$ epochs. Similarly, the network trained on the real object dataset was initially pre-trained on the synthetic dataset for $300$ epochs, and then fine-tuned for $500$ epochs on real object data. To determine the optimal parameters for training the network, an extensive exploration of various learning rates and optimizers was conducted through parameter sweeps. The results revealed that the combination of the ADAM optimizer~\cite{kingma2014adam} and a constant learning rate set at $10^{-5}$ performed optimally when using batches of $16$ point clouds. The implementation is based on PyTorch 1.6.0, and the network undergoes training on the Peregrine high-performance cluster, utilizing a computing node equipped with a single Nvidia V100 Tensor Core GPU. The training process spans 500 epochs, approximately 5 hours in total. During the preprocessing of each object's point cloud, a centering operation is applied, aligning the point cloud around the point $\operatorname{x}=0$, $\operatorname{y}=0$. The data is then normalized to fit within the range $[-1, 1]^3$. Throughout the training phase, point cloud augmentation is employed by introducing random noise from a normal distribution within the range $[-0.005, 0.005]$. This augmentation emulates sensor noise commonly encountered in real-world scenarios, enhancing the network's ability to generalize effectively to unseen data.

    % The results of training on the synthetic dataset and real-life dataset can be found in Fig.~\ref{fig:results}.
    % All plots are smoothed using a running average over the last 20 epochs for improved legibility. 

    By analyzing the accuracy versus epoch plot, we observed that both networks exhibited a rapid improvement in grasp accuracy, quickly convergent to adequate performance on the simulated dataset. The comparative results of grasping accuracy between the synthetic and real objects datasets were particularly interesting. On the synthetic dataset, both networks achieved roughly $94\%$ grasp accuracy, with GraspCaps outperforming the ablated network by less than $1\%$. However, when tested on the real object dataset, a substantial discrepancy emerged. Our approach demonstrated a $95\%$ grasp accuracy, whereas the ablated network converged to a significantly lower accuracy of $89\%$. This significant difference suggests that the GraspCaps architecture has a higher generalization ability than the ablated network. We hypothesize that the disparity in performance between the two networks on the real-world dataset can be attributed to the level of noise inherent in real-world data compared to the synthetic dataset. This robustness to noise and varying input data contributes to the superior performance of GraspCaps in real-world situations. This observation highlights the potential of the GraspCaps to adapt and generalize effectively, making it a promising solution for robotic applications.

\subsection{Simulation experiments }
\begin{figure}
    \centering
     \includegraphics[width =\linewidth]{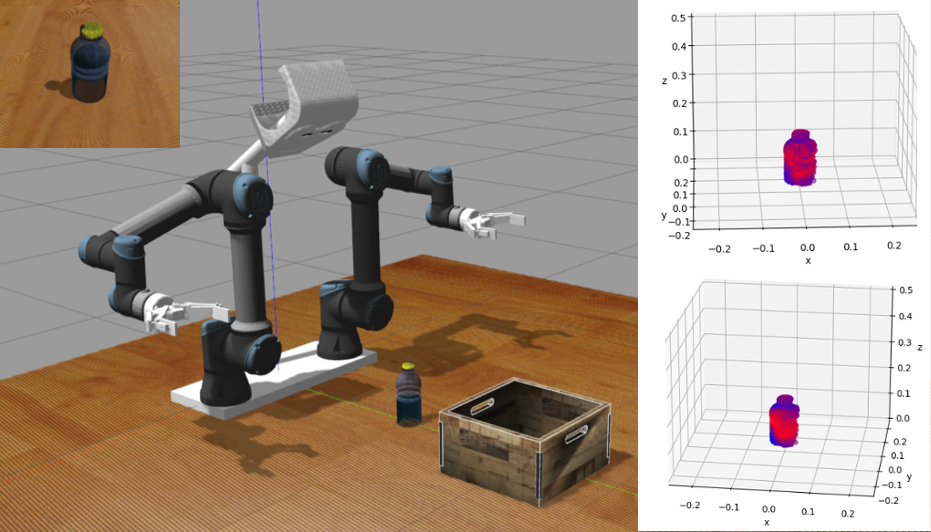}
        \caption{\textit{(left)} A bottle object is placed in front of the robot; \textit{(top-left)} what robot perceives through its camera; \textit{(top-right)} raw output of the fitness head of the GraspCaps network for the object; (\textit{lower-right}) the output when smoothed using our Gaussian filter.}
        \label{fig:smoothing_grasp}
       \vspace{-5mm}
\end{figure}

  To assess the performance of the proposed GraspCaps architecture in robotic grasping scenarios, we integrated it into our dual-arm robot setup within the Gazebo simulator. The robot is tasked with a multi-step process, starting with identifying the class of the object in its vicinity. Subsequently, the architecture generates a grasp configuration for picking the object. The robot then executes the grasp, lifting the object and placing it into the basket (see Fig.~\ref{fig:smoothing_grasp} - \textit{left}). Note that, we select the arm that is closest to the object to execute the grasp. This scenario is specifically designed to evaluate whether during the manipulation the object slips due to the suboptimal grasping configuration. The performance of the robot is quantified through the computation of a success rate, defined as $\frac{\#success}{\#attempts}$. In this set of experiments, a grasp is deemed successful if, by the experiment's conclusion, the object is securely placed inside the basket. This evaluation provides valuable insights into the practical applicability and success rate of GraspCaps in real-world robotic grasping scenarios.

% \begin{figure}
%     \centering
%         \begin{tabular}{cc}
%             \includegraphics[width = .45\linewidth]{images/bottle_before.png} &  
%             \includegraphics[width = .45\linewidth]{images/bottle_after.png}
%         \end{tabular}
%         \caption{Left: Raw output of the fitness head of the GraspCaps network on an object from the vertical cylinder class. Right: The output when smoothed using our Gaussian filter.}
%         \label{fig:smoothing_grasp}
%        \vspace{-5mm}
% \end{figure}

    To enhance the probability of accurately classifying objects in the input point set, we extract the full point set of the object, which often surpasses the required 1024 points. Multiple permutations of 1024 points are then generated and provided to the network. This approach mitigates the risk of the network misclassifying an object due to suboptimal point selection. The classification decision for the object is determined through a majority vote based on the neural network outputs. In cases of a tie, the algorithm iterates until a consensus is achieved. Post-processing of the network output involves smoothing the results from the fitness head of the network in the grasping module. This step is crucial for identifying regions with high fitness, indicative of suitable grasping locations. The smoothing process serves to refine the output, enhancing the network's ability to pinpoint areas conducive to successful grasping. The impact of this smoothing technique is exemplified in Fig.~\ref{fig:smoothing_grasp}. 
    % Such preprocessing and post-processing techniques contribute to the robustness and reliability of recognizing and accurately grasping diverse objects.
    We randomly present $10$ objects from each of the classes and instruct the robot to perform the pick-and-place scenario. We repeated this scenario five times to get a clear indication of the performance of both networks. We found that, although both the GraspCaps network and the ablated network achieved a similar classification performance of $86\%$, the GraspCaps network outperformed the ablated network in terms of grasping success with percentages of $71\%$ for the GraspCaps network versus $53\%$ for the ablated network. Following a thorough investigation of the results, we identified the cutlery class as a prominent negative outlier, exhibiting suboptimal performance in both classification accuracy (30\%, compared to the average accuracy of 86\%) and grasping success (50\%, compared to the average success of 71\%). To uncover the underlying cause of this discrepancy, we conducted an in-depth analysis using the confusion matrix generated from GraspCaps classifications. Our analysis showed a consistent pattern: all misclassifications within the cutlery class were consistently assigned to the category of vertical objects. The primary challenge stems from the fact that vertical objects are typically grasped in a sideways orientation, which contrasts with the top-down grasps more suitable for objects in the cutlery class. Notably, cutlery items are characterized by their thin profile, often resulting in a point cloud containing fewer than 500 points. The visibility of these objects varies based on their pose relative to the camera and self-occlusion. We therefore upsampled the point cloud of the object to reach the required 1024 points. However, this process introduced changes to the object's shape. In some instances, this transformation made the object more closely resemble a vertical cylinder, contributing to confusion in the classification process and explaining the observed poor grasping performance. Mitigating this challenge involves adopting shape completion techniques~\cite{Cheng_2023_CVPR,Li_2023_CVPR,delgado2023shape} that are more robust to variations in the shape of thin objects like cutlery items.

    % See Tables~\ref{tab:graspcaps_synth} and \ref{tab:combined_synth} for for a per-class breakdown of the achieved classification and grasping accuracy of the two networks. 
    % A short video demonstrating the performance of GraspCaps has been attached to the paper as supplementary material.
    % can be found on online at \texttt{https://youtu.be/89A1gzKJGxI}
    % In Fig.~\ref{fig:conf_mat} a confusion matrix of the classification performance of the GraspCaps network can be found.
    
    %When looking at the performance of the GraspCaps architecture in Table~\ref{tab:graspcaps_synth}, it can be seen that the cutlery class is a clear negative outlier in terms of both classification accuracy (30\% vs. the average accuracy of 86\%) and grasping success (50\% vs. the average success of 71\%). When looking at the confusion matrix of the GraspCaps classifications in Fig.~\ref{fig:conf_mat}, we observe that all misclassifications of the cutlery class were classified as vertical objects. These are often grasped in a sideways fashion, as opposed to the top-down grasps that should be employed when grasping objects from the cutlery or horizontal cylinder classes, which explains the poor grasping performance. 
    
\subsection{Grasping novel objects experiments}
\begin{figure}[!t]
    \centering
         \includegraphics[width =\linewidth, trim = 0cm 0cm 0cm 0cm clip=true]{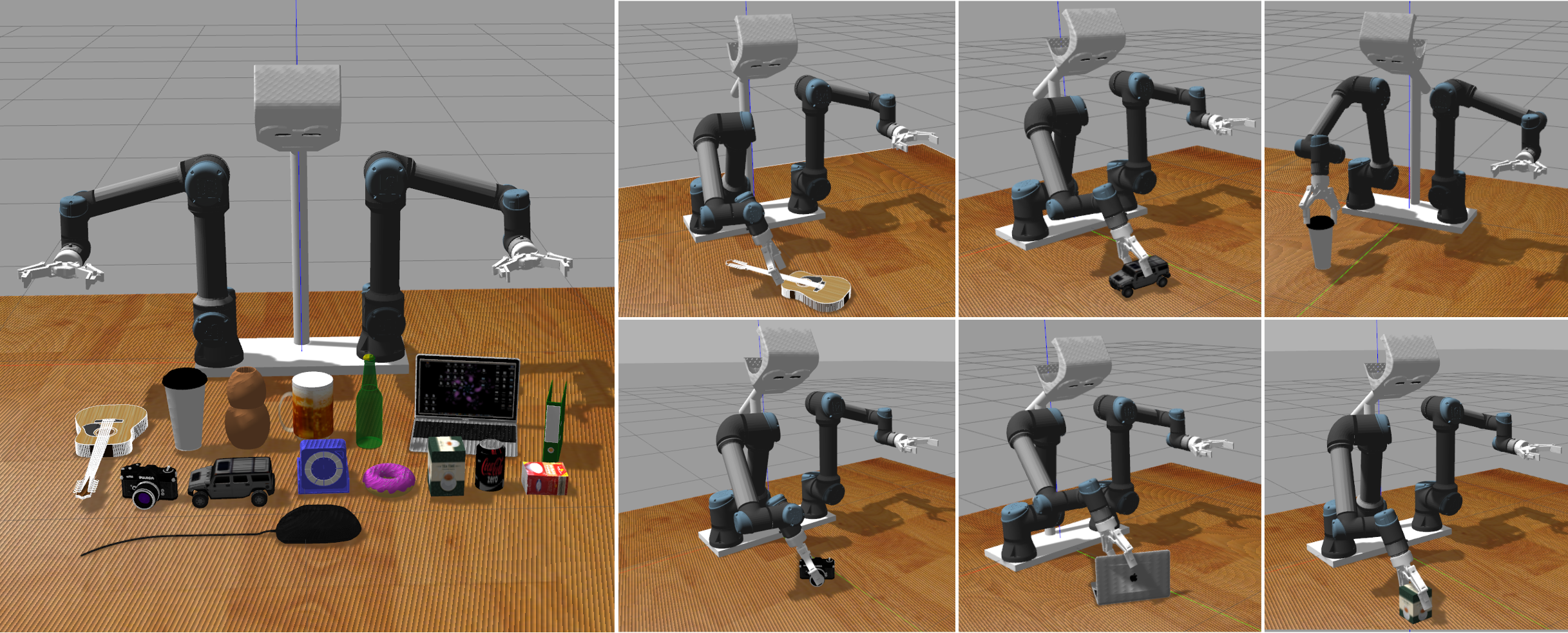}\\
         \includegraphics[width =\linewidth]{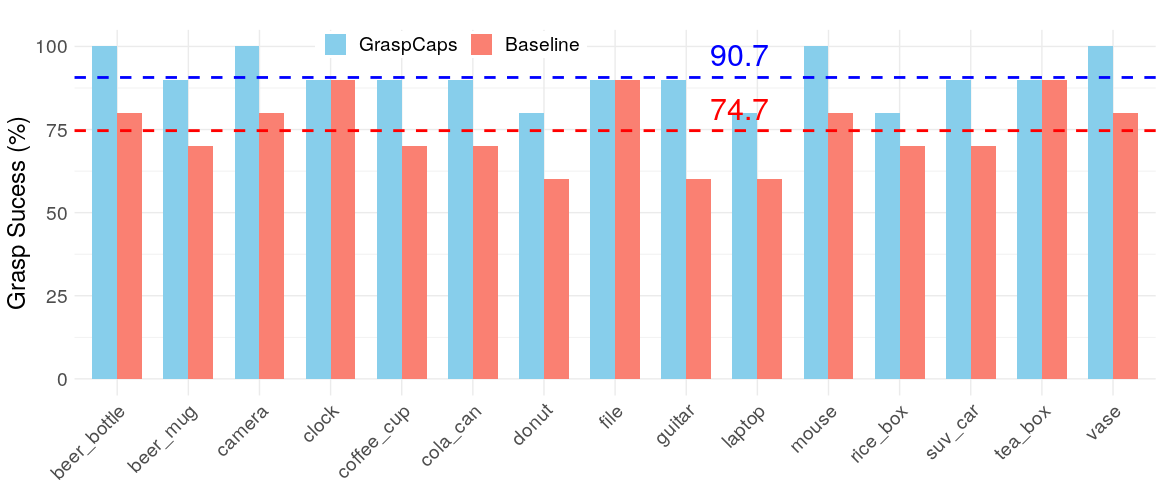}
         \vspace{-6mm}
        \caption{(\textit{Top-left}) The collection of novel objects; (\textit{top-right}) demonstration of how the robot grasped various objects using GraspCaps; (\textit{lower-row}) results of grasping novel objects experiments for GraspCaps and the baseline networks. The dashed lines represent the average success rate achieved by the models.}
        \label{fig:sim_results}
       \vspace{-5mm}
\end{figure}

To assess the generalizability of the network, we conducted similar pick-and-place tests on a set of 15 novel objects, retrieved from the 3DGEMS dataset~\cite{rasouli2017effect} and Gazebo repository. These objects are shown in Fig.~\ref{fig:sim_results} (\textit{top-left}). 
% The experimental setup and task will be equal to the setup in Gazebo described earlier in this section. 

A number of the novel objects, such as the `\textit{bear-bottle}', `\textit{vase}', and `\textit{tea-box}', can be sorted into the classes the network was trained on. Other objects, such as the laptop, guitar, mouse, camera, and digital clock, have more complex shapes and may be difficult to sort into the existing classes. Therefore, it will be interesting to see how the network will adapt and if it will be able to successfully grasp the objects. Both the GraspCaps architecture and the ablated architecture will attempt to grasp each of the objects $10$ times to get a clear indication of the generalization capability of the networks. Results are summarized in Fig.~\ref{fig:sim_results} (\textit{lower-row}). When comparing the results of the GraspCaps network with the baseline network, it can be seen that the GraspCaps network outperforms the baseline network with an average of $90.7\%$ grasping accuracy for the GraspCaps architecture compared to an average of $74.7\%$ for the baseline network. We observed that for familiar classes like `\textit{clock}', `\textit{file}', and `\textit{tea-box}', both baseline and GraspCaps perform well with $90\%$ success rate. However, for unfamiliar classes (e.g., `\textit{laptop}', `\textit{donut}', `\textit{camera}', `\textit{vase}', etc.), the part-whole properties inherent in capsule networks enable GraspCaps to outperform the baseline approach. GraspCaps consistently outperforms the Baseline across the remaining categories. This suggests that GraspCaps excels in grasping objects, potentially due to its ability to consider finer details in the object's geometry. We observed that most of the failures were mainly due to predicting the grasp point near the edge of an object. As the gripper initiated the closing motion, the object's position could be affected, leading to potential displacement. It is crucial to emphasize that even minor camera-robot calibration errors can significantly exacerbate these issues, ultimately hindering the robot's ability to successfully grasp the target object. The refinement of predicted grasp poses falls outside the scope of this work and has been addressed in existing literature~\cite{mousavian20196, urain2023se, wei2021gpr}.

% \begin{figure}[!t]
%     \centering
%          \includegraphics[width =\linewidth]{images/graspcaps_baseline_2.png}
%          \vspace{-8mm}
%         \caption{Results of grasping novel objects experiments for GraspCaps and the baseline networks. The dashed lines represent the average grasp success rate achieved by each of the methods.}
%         \label{fig:sim_novel_object_results}
%        \vspace{-0mm}
% \end{figure}

\subsection{Real-robot experiments}

\begin{figure}[!t]
        \centering
        \includegraphics[width=\linewidth]{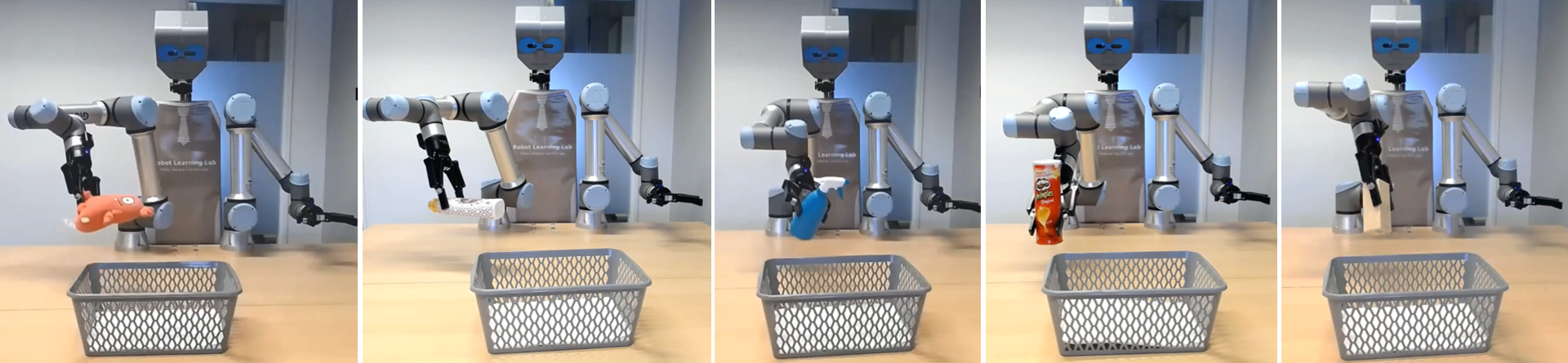}\\
        \includegraphics[width=\linewidth]{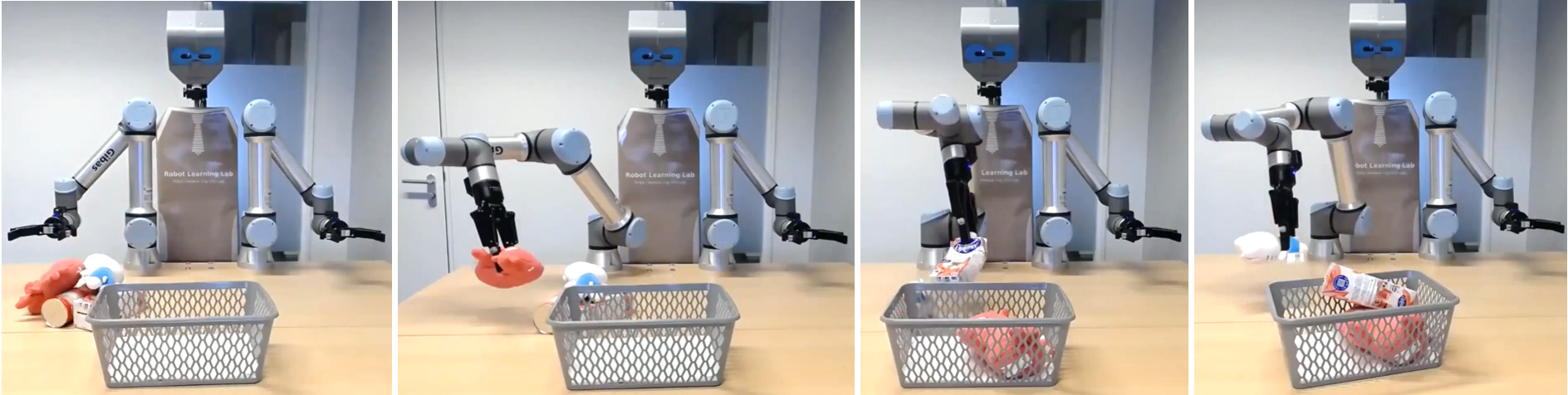}
        \vspace{-5mm}
        \caption{Real robot experiments: (\textit{top-row}) isolated object scenarios; (\textit{lower-row}) a pile of objects scenario.}
        \label{fig:real_robot_exps}
        \vspace{-5mm}
\end{figure}

In this set of experiments, we conducted pick-and-place tasks using a real dual-arm robot~\cite{kasaei2018towards}\cite{kasaei2019interactive}\cite{kasaei2021simultaneous} to assess the effectiveness of the proposed approach in both isolated object and pile of objects scenarios. The GraspCaps network underwent testing in a real-world environment and demonstrated robust performance not only on objects present in its dataset but also on novel objects. Fig.~\ref{fig:real_robot_exps} showcases five instances of grasping objects in isolated scenarios, along with a sequence of snapshots illustrating a pile experiment.
    
In the isolated object scenarios, the GraspCaps network consistently exhibited accurate and reliable performance, efficiently grasping objects and placing them into the basket in the majority of trials. These experiments included a variety of objects, some of which were not present in the training dataset. The successful execution of pick-and-place tasks in this setting demonstrates the network's adaptability to generalize to novel objects and competence in manipulating diverse objects. In the pile of objects scenario, five objects were placed with random rotations and significant overlap on top of each other resulting in some objects being partially or fully obscured from view. The GraspCaps network demonstrated good performance in handling such cases. Despite the challenges posed by obscured views, the network adeptly navigated the cluttered environment, successfully grasping objects from the pile and dropping them into the basket. This experiment showed that GraspCaps does not rely on pre-segmented point clouds and can handle complex scenarios with multiple objects and occlusions. These results highlight the practical applicability and adaptability of the GraspCaps architecture in real-world robotic manipulation tasks, even in complex and cluttered environments. To further illustrate the network's performance, a supplementary video has been included with the paper: \href{https://youtu.be/d13rEhKgApI?si=EhgbDI84nlXL5V2M}{https://youtu.be/d13rEhKgApI?si=EhgbDI84nlXL5V2M}. 
% This video provides a visual demonstration of the GraspCaps network in action, showcasing both successful and unsuccessful attempts in both isolated object and pile scenarios. Moreover, the video features instances with novel objects, further emphasizing its adaptability to unfamiliar objects and real-world challenges.

%% file: sec/conclusion.tex
\section{Conclusions}
\label{sec:conclusion}
    % This work introduced GraspCaps, an innovative model designed for simultaneous recognition and grasping, utilizing capsule networks.  Along with GraspCaps we have presented a new dataset for 6DoF grasping based on point cloud data. Our experiments demonstrated that GraspCaps performed well in both simulated and real-world scenarios across a diverse range of familiar and novel objects.     As a direction for future research, we are interested in extending the network by incorporating an additional head that generates an affordance mask. This enhancement could enable the model to focus on specific regions of objects, facilitating task-informed grasping, such as grasping handles of a pan or the handle of a knife. This extension would contribute to the adaptability and versatility of GraspCaps in real-world applications, where precise grasping strategies are crucial for successful interactions with various objects.

This work introduced GraspCaps, an innovative model designed for familiar 6D object grasping based on capsule networks. Along with GraspCaps, we presented a new dataset for 6DoF grasping based on point cloud data. To validate the performance of our approach, we conducted sets of experiments in simulation and with a real robot. Our experiments demonstrated that GraspCaps performed well in both simulated and real-world scenarios across a diverse range of objects. However, it is essential to acknowledge the limitations of this work, particularly in instances where precise prediction of grasp points for thin objects proved challenging. Further research is needed to address these limitations and enhance the robustness of GraspCaps. For future research, we are interested in extending the network by incorporating an additional head that generates an affordance mask. This enhancement could enable the model to focus on specific regions of objects, facilitating task-informed grasping, such as grasping handles of a pan or the handle of a knife. This extension would contribute to the versatility of GraspCaps in real-world applications, where precise grasping is crucial for safe interactions with various objects.

%% file: sec/ack.tex
\section*{Acknowledgement}
We thank the Center for Information Technology of the University of Groningen for their support and for providing access to the Peregrine high-performance computing cluster.